\documentclass[conference]{IEEEtran}
\IEEEoverridecommandlockouts
\usepackage{cite}
\usepackage{amsmath,amssymb,amsfonts}
\usepackage{algorithmic}
\usepackage{graphicx}
\usepackage{textcomp}
\usepackage{xcolor}
\usepackage{hyperref}

\usepackage{booktabs}
\usepackage{diagbox}
\usepackage{multirow}

\usepackage{placeins}
\usepackage{float}

\begin{document}

\title{LLMs Meet Finance: Fine-Tuning Foundation Models for the Open FinLLM Leaderboard\\

\thanks{\textsuperscript{\dag}These authors contributed equally to this work.}
\thanks{\textsuperscript{*}Corresponding author: Haizhao Yang \href{mailto:hzyang@umd.edu}{hzyang@umd.edu}.}
}

\author{\IEEEauthorblockN{1\textsuperscript{st} Varun Rao\textsuperscript{\dag}}
\IEEEauthorblockA{
\textit{University of Maryland, College Park}\\
MD, 20742\\
vrao1234@terpmail.umd.edu
}
\and
\IEEEauthorblockN{2\textsuperscript{nd} Youran Sun\textsuperscript{\dag}}
\IEEEauthorblockA{
\textit{Department of Mathematics}\\
\textit{University of Maryland, College Park}\\
MD, 20742 \\
syouran0508@gmail.com}
\and
\IEEEauthorblockN{3\textsuperscript{rd} Mahendra Kumar}
\IEEEauthorblockA{
\textit{University of Maryland, College Park}\\
MD, 20742\\
mahen037@terpmail.umd.edu
}
\and
\IEEEauthorblockN{4\textsuperscript{th} Tejas Mutneja}
\IEEEauthorblockA{
\textit{University of Maryland, College Park}\\
MD, 20742\\
tmutneja@terpmail.umd.edu
}
\and
\IEEEauthorblockN{5\textsuperscript{th} Agastya Mukherjee}
\IEEEauthorblockA{
\textit{University of Maryland, College Park}\\
MD, 20742\\
amukhs13@terpmail.umd.edu
}
\and
\IEEEauthorblockN{6\textsuperscript{th} Haizhao Yang\textsuperscript{*}}
\IEEEauthorblockA{
\textit{Department of Mathematics} \\
\textit{Department of Computer Science} \\
\textit{University of Maryland, College Park}\\
MD, 20742 \\
hzyang@umd.edu}
}

\maketitle

\begin{abstract}
This paper investigates the application of large language models (LLMs) to financial tasks.
We fine-tuned foundation models using the Open FinLLM Leaderboard as a benchmark.
Building on Qwen2.5 and Deepseek-R1, we employed techniques including supervised fine-tuning (SFT), direct preference optimization (DPO), and reinforcement learning (RL) to enhance their financial capabilities.
The fine-tuned models demonstrated substantial performance gains across a wide range of financial tasks.
Moreover, we measured the data scaling law in the financial domain.
Our work demonstrates the potential of large language models (LLMs) in financial applications.
\end{abstract}






\section{Introduction}

With the development of Transformer-based Natural Language Processing (NLP), Large Language Models (LLMs) are becoming a standard technology integrated into the human toolkit.
In the financial sector, LLMs are expected to assist in tasks such as making intelligent decisions and creating personalized financial searches.
The integration of LLMs into financial applications is expected to advance the development of open finance.
This paper aims to explore the applications of reinforcement learning (RL) and LLMs in financial tasks.

The Open FinLLM Leaderboard is an open platform for evaluating the performance of LLMs on various financial tasks.
It is based on Finben (Pixiu) \cite{finben2024} and includes 36 datasets spanning 24 financial tasks, covering seven critical aspects: information extraction (IE), textual analysis, question answering (QA), text generation, risk management, forecasting, and decision-making.
In this paper, we employ various techniques, including prompt engineering, chain-of-thought (CoT) reasoning, and reinforcement learning (RL), to enhance the performance of open-source LLMs on the Open FinLLM Leaderboard.

Our main contributions are as follows:  
\begin{enumerate}
    \item We fine-tuned the Qwen2.5 and Deepseek-R1 models using techniques such as supervised fine-tuning (SFT) and direct preference optimization (DPO) and observed significant improvements in the models' performance across diverse tasks.
    \item We leverage reinforcement learning (RL) and chain-of-thought (CoT) prompting to synthesize data and further strengthen the capabilities of LLMs in the financial domain.
    \item We measured the data scaling law in the financial domain and found it highly consistent with results reported in previous literature.
\end{enumerate}

\section{Related Work}

\paragraph{Financial Reinforcement Learning Datasets and Frameworks}
FNSPID \cite{FNSPID} provides a large-scale dataset spanning from 1999 to 2023, containing 29.7 million stock price records and 15.7 million time-aligned financial news articles for 4,775 S\&P 500 companies. FinRL-Meta \cite{FinRL-Meta, DynamicDatasets} is a data-centric library for financial reinforcement learning that transforms dynamic real-world market data into gym-style environments.

\paragraph{LLM-Enhanced Agents}
FinRL-DeepSeek \cite{FinRL-DeepSeek} introduces a trading agent that combines reinforcement learning with large language models to enhance trading performance. \cite{reportfinancialregulationschallenge} summarized the approaches and results of participating teams in the Financial Regulations Challenge at COLING 2025.
FinMind-Y-Me \cite{FinMind-Y-Me} ranked first in the competition mentioned above. In their solution, they applied techniques such as Sequential Fine-Tuning and Task-Specific Prompts to enhance the performance of the fine-tuned model.

\paragraph{Simulate micro-level behaviors using LLMs} Some studies use LLM-based multi-agent systems to simulate micro-level behaviors, aiming to observe financial market dynamics and macroeconomics after removing the assumption of fully rational agents \cite{simulatingfinancial2024, econagent2024}.

\begin{table*}[ht]
    \centering
    \caption{Evaluation of DeepSeek and Qwen on multiple tasks before and after fine-tuning.}
    \begin{tabular}{ccc|cccc}
\toprule
Model & Fine-tuned on &Datasize (\#Tokens)& NER (F1) & CC (F1) & FiQASA (F1) & FPB (F1)\\
\midrule
\multirow{6}{*}{DeepSeek-R1-1.5B} & / & / & 0.1448 & \textbf{0.6683} & 0.4383 & 0.1845\\
& NER & 408 (51k) & \textbf{0.7231} & 0.3290 & 0.2755 & 0.1433\\
& FiQASA & 750 (34k) & 0.0286 & 0.1977 & 0.7865 & \textbf{0.2123}\\
& NER\&FiQASA & 1158 (85k) & 0.6913 & 0.0950 & 0.7584 & 0.2019\\
& NER, then FiQASA & 1158 (85k) & 0.7002 & 0.0967 & \textbf{0.7866} & 0.1355\\
& FiQASA, then NER & 1158 (85k) & 0.6540 & 0.0822 & 0.6984 & 0.1461\\
\midrule
\multirow{6}{*}{Qwen2.5-1.5B-Instruct} & / & / & 0.0060 & \textbf{0.6667} & 0.6789 & 0.2269\\
& NER & 408 (51k) & \textbf{0.6212} & 0.3962 & 0.6894 & 0.2141\\
& FiQASA & 750 (34k) & 0.0000 & 0.5144 & 0.8029 & 0.5016\\
& NER\&FiQASA & 1158 (85k) & 0.5926 & 0.5357 & 0.7703 & \textbf{0.6704}\\
& NER, then FiQASA & 1158 (85k) & 0.0938 & 0.3055 & \textbf{0.8054} & 0.2359\\
& FiQASA, then NER & 1158 (85k) & 0.4873 & 0.2064 & 0.7999 & 0.3622\\
\bottomrule
    \end{tabular}\\
    \vspace{1ex}
\noindent Beyond reporting scores on the fine-tuned task, we include results on other tasks to investigate the models' generalization and transfer learning capabilities. Task abbreviations: NER (Named Entity Recognition), CC (Causal Classification), FiQASA (Financial QA-Style Sentiment Analysis), FPB (Financial Polarity Benchmark).
    \label{tab:main_result1}
\end{table*}

\section{Methodology}

\paragraph{Supervised Fine-Tuning (SFT)} SFT is mainly used for alignment, task adaptation, and knowledge enhancement.
Therefore, we always start with SFT.
Among all 41 datasets, 28 provide training/validation data, and we use these data for SFT.
Formally, the loss is given by
\[
L_{\textrm{SFT}} = - \ln \pi_{\theta} (y^+ | x)
\]
where $\theta$ represents the parameters of the LLM, $x$ is the query, and $y^+$ is the answer.

Apart from directly applying SFT on the training set, performance can be further improved by employing Sequential Fine-Tuning and Task-Specific Prompts, as mentioned in FinMind-Y-Me \cite{FinMind-Y-Me}.

\paragraph{Direct Preference Optimization (DPO)} After SFT, the performance of the large model improved significantly.
However, we observed that the LLM tends to generate excessively long and repeating responses \cite{repeater2020,repeater2025}.
For example, if the correct answer is ``Apple", the model might output ``Apple Apple Apple ..." due to its inability to determine a proper stopping point..
To ensure that the model stops at the appropriate point, we apply DPO fine-tuning after SFT. Specifically, the DPO loss is defined as
\[
L_{\text{DPO}} = -\ln \sigma \left( \beta \cdot \left( \ln \frac{\pi_{\theta}(y^+ | x)}{\pi_{\text{ref}}(y^+ | x)} - \ln \frac{\pi_{\theta}(y^- | x)}{\pi_{\text{ref}}(y^- | x)} \right) \right)
\]
where \( \pi_{\theta} \) represents the current model, \( \pi_{\text{ref}} \) is the base model after SFT, \( \beta \) is a temperature hyperparameter that controls the optimization strength (usually set to 1) and \( \sigma(\cdot) \) is the Sigmoid function.
For the positive data \( y^+ \), we use the answers from the database, while the negative data \( y^- \) consists of overly long responses (or outputs that we consider undesirable).

\paragraph{Data Synthesis and Reinforcement Learning (RL)}
In some cases, the dataset does not contain training data, and we have to synthesize data for reinforcement learning. The specific procedure for RL is as follows:  
(1) Corpus Collection: gather relevant text data based on the theme of the dataset.  
(2) LLM Annotation: the current LLM \( \pi_t \) is prompted to generate annotations using Chain-of-Thought (CoT) reasoning.  
(3) Answer Extraction: apply regular expressions to extract the answer \( y^+ \) from the generated response.  
(4) Query Formatting: the extracted data is structured into a query \( x \) following the prompt format of the dataset.  
(5) Training: use the synthesized \( x, y^+ \) pairs for SFT and DPO to train and denote the result LLM as \( \pi_{t+1} \).  

If computational resources allow, we can further use \( \pi_{t+1} \) to generate additional data, then continue SFT or DPO to obtain \( \pi_{t+2} \). This iterative process can be repeated multiple times to refine the model.

\begin{figure}[ht]
    \centering
    \includegraphics[width=0.95\linewidth, trim=0 5 0 0, clip]{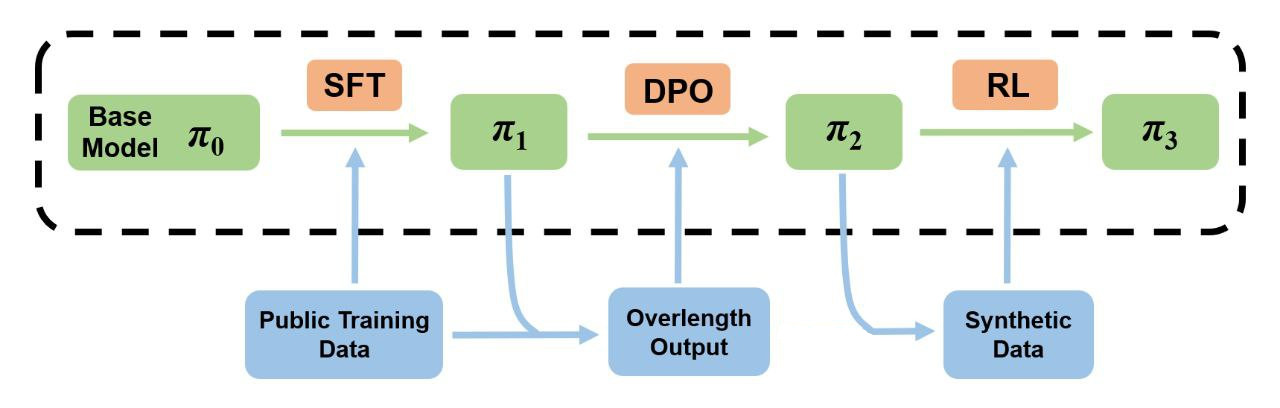}
    \caption{Training flowchart showing the progression from base model to final model through SFT, DPO, and RL stages.}
    \label{fig:pipeline}
\end{figure}

The training flowchart is illustrated in Figure \ref{fig:pipeline}.

\section{Experiments}

\paragraph{The effectiveness of SFT}\label{para:sft}
We fine-tuned the models using LLaMAFactory \cite{LlamaFactory2024} with a learning rate of 5e-5 and a constant learning rate scheduler.
The effective batch size was set to 64.
We used LoRA with a rank of 128 and an alpha of 256.
DeepSpeed \cite{DeepSpeed2023} stage 2 was applied during training.

Table \ref{tab:main_result1} presents the results of training DeepSeek-R1-1.5B and Qwen2.5-1.5B-Instruct using data from NER (Named Entity Recognition) and FiQASA (Financial QA-Style Sentiment Analysis).
Beyond reporting scores on the fine-tuned tasks, we also include results on CC (Causal Classification) and FPB (Financial Polarity Benchmark) to investigate the models’ generalization and transfer learning capabilities.
We can observe that SFT significantly improves performance in the corresponding task.
DeepSeek-R1-1.5B achieves higher performance metrics than Qwen2.5-1.5B-Instruct, indicating better task-specific generalization.

However, the interactions between different tasks do not follow a general pattern.
Training on NER reduces the performance on CC, while training on FiQASA enhances the performance on FPB.
Both the FiQASA and FPB datasets involve sentiment classification tasks over sets of headlines. The similarity in task structure and headline formatting between these datasets likely accounts for the substantial improvement in performance on FPB benchmarks following initial fine-tuning on FiQASA.
In contrast, the NER dataset, which focuses primarily on the identification of named entities, may have hindered performance on the CC benchmark. This could be explained by the fine-tuned model overemphasizing named entities at the expense of capturing the underlying semantic or causal relationships between headlines.
Moreover, Sequential Fine-Tuning does not outperform direct fine-tuning across the tasks.

\begin{table}[ht]
    \centering
    \caption{Effect of DPO after SFT.}
    \begin{tabular}{c|ccc}
    \toprule
    Train Phase & Overlength Ratio & NER & CC\\
    \midrule
    SFT & 54.7\% & 0.6212 & 0.3962\\
    +1 DPO & 8.6\% & 0.6180 & 0.4437 \\
    +1 DPO & \textbf{1.7\%} & 0.5950 & \textbf{0.5592}\\
    \bottomrule
    \end{tabular}\\
    \vspace{1ex}
The Overlength Ratio refers to the proportion of samples in the training set where the number of output tokens exceeds the number of answer tokens.
    \label{tab:dpo_result}
\end{table}

\paragraph{The effectiveness of DPO}
As discussed in the methodology section, after SFT, the model tends to produce repetitive outputs; in other words, it learns to ``repeat the correct answer tokens" rather than understanding the relationship between the correct answer and the context.
In this task, we are unable to modify the prompts or change the sampling parameters \cite{repeater2020}, so we apply DPO after SFT.
We treat overly long outputs generated by the SFT model on the training set as rejected labels and use the correct answers as accepted labels.

Table \ref{tab:dpo_result} presents the results of our DPO experiments.
We first perform SFT on Qwen2.5-1.5B-Instruct using an NER dataset and then apply DPO based on the fine-tuned model.
The SFT configuration is the same as in Table \ref{tab:main_result1}. For DPO, the training setup is identical to that of SFT, except that the LoRA rank and LoRA alpha are reduced to 16 and 32, respectively.
As shown, the Overlength Ratio is significantly reduced after applying DPO, which aligns with our original motivation for using DPO.
In addition to the Overlength Ratio, we also recorded the F1 scores for the NER and CC tasks to investigate the impact of DPO on other tasks.
We observe that the F1 score for the trained NER task remains unchanged mainly (the slight decrease may be due to the scoring mechanism of the benchmark), while the score for the unseen CC task actually improves.

\begin{table}[ht]
    \centering
    \caption{The effectiveness of using synthetic data for RL.}
    \begin{tabular}{cccc}
    \toprule
        Task & Metrics & Datasize (\#Tokens) & Performance Boost \\
    \midrule
       MultiFin & F1 & 219 (23k) & +87.1\% \\
       FOMC & F1 & 108 (10k) & +22.5\% \\
       TSA & RMSE & 105 (10k) & +3.4\% \\
    \bottomrule
    \end{tabular}\\
    \vspace{1ex}
    Task abbreviations (description): MultiFin (Real-world Article Headlines across different writing systems and language families), FOMC (Federal Open Market Committee Hawkish-Dovish Classification), TSA (Sentiment Analysis on Social Media).
    \label{tab:rl}
\end{table}

\paragraph{RL with Synthesed Data}

Almost half of the tasks do not provide training datasets, so we need to synthesize data to improve the model's performance on these tasks.
The data synthesis process consists of three steps:  
(1) Corpus Collection: Prompt the LLM to generate questions based on the task;  
(2) LLM Annotation: Generate annotations using Chain-of-Thought (CoT) reasoning;  
(3) Answer Extraction: Use regular expressions to extract the answer $y^{+}$ from the generated response.
We then apply the aforementioned method for fine-tuning.

Table \ref{tab:rl} presents the results of training for one iteration using RL on three different tasks.
We used DeepSeek-R1-1.5B to synthesize data, which was subsequently used to train the same model.
The SFT parameters remain consistent with those presented in Table \ref{tab:main_result1}.
We observe performance improvements across all tasks.

\paragraph{Data Scaling Law}

\begin{figure}[h]
    \centering
    \includegraphics[width=0.95\linewidth]{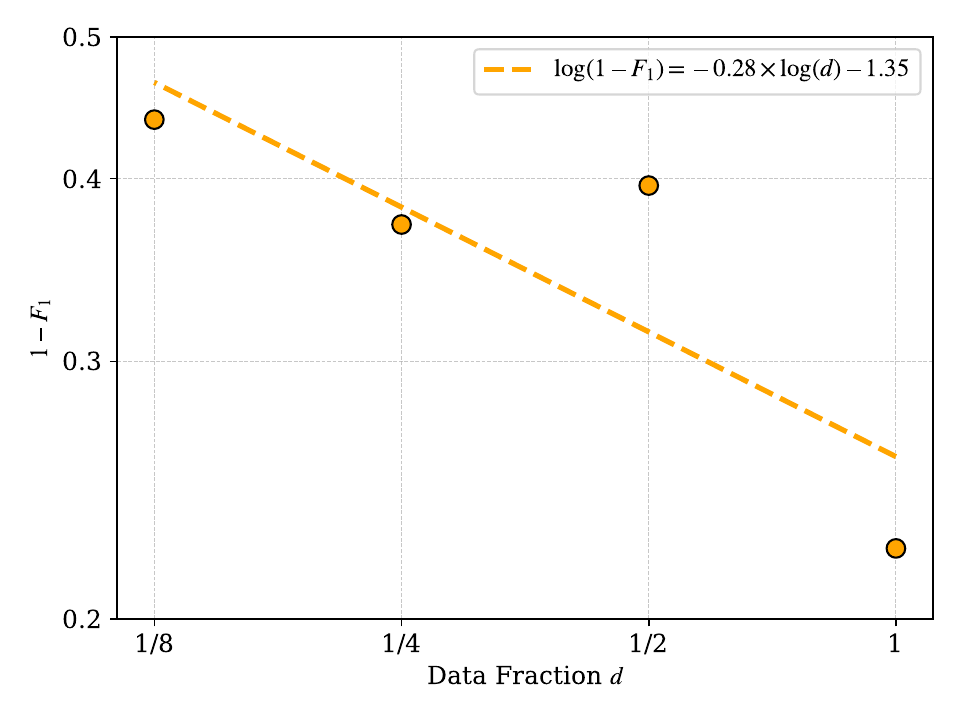}
    \vspace{-2ex}
    \caption{The data scaling law on financial tasks.}
    \label{fig:scaling}
\end{figure}

An essential property of large language models is the presence of scaling laws, which allow us to predict the performance of larger-scale models based on the results of a small number of smaller-scale experiments.
To investigate the data scaling behavior on financial tasks, we aggregated all available data and conducted fine-tuning with varying fractions of the dataset (100\%, 50\%, 25\%, and 12.5\%).
The average F1 scores obtained from each setting are presented in Figure \ref{fig:scaling}.

It is worth noting that the data critical exponent we measured is highly consistent with results reported in previous literature. Specifically, \cite{openai2020} reports the following scaling laws
$
    L \sim p^{-0.076}, \; L \sim d^{-0.095},
$
where \( L \) denotes the test loss, \( p \) the number of parameters, and \( d \) the dataset size.
A relation of the form \( x \sim y^{\alpha} \) corresponds to \( \log(x) = \alpha \log(y) + C \) for some constant \( C \), and the exponent \( \alpha \) is referred to as the \textbf{critical exponent}.
On the other hand, \cite{tsinghua2024} reports
$
    E \sim p^{-0.195},
$
where $E$ is the test error rate.
By combining these results, it is straightforward that
\begin{equation}
    E\sim L^{0.195/0.076=2.57}\sim d^{\,2.57\times0.095=0.24}.
\end{equation}
Our measured exponent of $\mathbf{0.28}$ is very close to this derived value of $\mathbf{0.24}$, which not only supports the correctness of our experiments, but also suggests a broader implication: while the results in \cite{openai2020} are obtained on general text and those in \cite{tsinghua2024} on math, our findings are based on financial tasks. 
This consistency across domains implies that the critical exponent may be independent of tasks; in other words, it exhibits a form of universality \cite{phasetransitions}.

\paragraph{Computational Cost}

The tasks require the use of high-performance GPUs.
Fine-tuning each 1.5B model required approximately 3 to 5 hours on 4x2080Ti.
Additionally, generating new training sets for specific datasets without publicly available training sets took an additional 1 to 2 hours per dataset.

\section{Discussion and Prospect}

\paragraph{Preprocessing and Postprocessing} In this competition, we can only upload models and cannot make modifications to preprocessing (prompt engineering) or postprocessing (generation configuration and parsing of output). However, in practice, preprocessing and postprocessing are often more convenient and can be as effective as fine-tuning in improving model performance. 
For example, the issue of DeepSeek models receiving lower scores due to the output of chain-of-thought (CoT) reasoning can be addressed by adding more examples and instructions in the prompts or by using regular expressions in postprocessing to remove the CoT process.
Another example is the repeater phenomenon after model fine-tuning, which is often mitigated by increasing the repeat penalty or using top\_p (nucleus sampling) instead of top\_k.
We believe that if the organizers grant us the flexibility to apply preprocessing and postprocessing, we can further enhance the model's performance.

\paragraph{Prospect}
Due to time constraints, many potentially beneficial experiments were left unexplored:
(1) Testing larger models;  
(2) Exploring task-specific prompts to improve performance on reasoning-intensive tasks further;
(3) Running additional iterations of RL training to examine whether performance saturates.

\section*{Acknowledgment}

The authors were partially supported by the US National Science Foundation under awards DMS-2244988, DMS2206333, the Office of Naval Research Award N00014-23-1-2007, and the DARPA D24AP00325-00.

\bibliographystyle{IEEEtran}
\bibliography{main}

\begin{thebibliography}{10}
\providecommand{\url}[1]{#1}
\csname url@samestyle\endcsname
\providecommand{\newblock}{\relax}
\providecommand{\bibinfo}[2]{#2}
\providecommand{\BIBentrySTDinterwordspacing}{\spaceskip=0pt\relax}
\providecommand{\BIBentryALTinterwordstretchfactor}{4}
\providecommand{\BIBentryALTinterwordspacing}{\spaceskip=\fontdimen2\font plus
\BIBentryALTinterwordstretchfactor\fontdimen3\font minus \fontdimen4\font\relax}
\providecommand{\BIBforeignlanguage}[2]{{%
\expandafter\ifx\csname l@#1\endcsname\relax
\typeout{** WARNING: IEEEtran.bst: No hyphenation pattern has been}%
\typeout{** loaded for the language `#1'. Using the pattern for}%
\typeout{** the default language instead.}%
\else
\language=\csname l@#1\endcsname
\fi
#2}}
\providecommand{\BIBdecl}{\relax}
\BIBdecl

\bibitem{finben2024}
\BIBentryALTinterwordspacing
Q.~Xie, W.~Han, Z.~Chen, R.~Xiang, X.~Zhang, Y.~He, M.~Xiao, D.~Li, Y.~Dai, D.~Feng, Y.~Xu, H.~Kang, Z.~Kuang, C.~Yuan, K.~Yang, Z.~Luo, T.~Zhang, Z.~Liu, G.~Xiong, Z.~Deng, Y.~Jiang, Z.~Yao, H.~Li, Y.~Yu, G.~Hu, J.~Huang, X.-Y. Liu, A.~Lopez-Lira, B.~Wang, Y.~Lai, H.~Wang, M.~Peng, S.~Ananiadou, and J.~Huang, ``Finben: A holistic financial benchmark for large language models,'' 2024. [Online]. Available: \url{https://arxiv.org/abs/2402.12659}
\BIBentrySTDinterwordspacing

\bibitem{FNSPID}
\BIBentryALTinterwordspacing
Z.~Dong, X.~Fan, and Z.~Peng, ``Fnspid: A comprehensive financial news dataset in time series,'' in \emph{Proceedings of the 30th ACM SIGKDD Conference on Knowledge Discovery and Data Mining}, ser. KDD '24.\hskip 1em plus 0.5em minus 0.4em\relax New York, NY, USA: Association for Computing Machinery, 2024, p. 4918–4927. [Online]. Available: \url{https://doi.org/10.1145/3637528.3671629}
\BIBentrySTDinterwordspacing

\bibitem{FinRL-Meta}
\BIBentryALTinterwordspacing
X.-Y. Liu, Z.~Xia, J.~Rui, J.~Gao, H.~Yang, M.~Zhu, C.~D. Wang, Z.~Wang, and J.~Guo, ``Finrl-meta: Market environments and benchmarks for data-driven financial reinforcement learning,'' 2022. [Online]. Available: \url{https://arxiv.org/abs/2211.03107}
\BIBentrySTDinterwordspacing

\bibitem{DynamicDatasets}
\BIBentryALTinterwordspacing
X.-Y. Liu, Z.~Xia, H.~Yang, J.~Gao, D.~Zha, M.~Zhu, C.~D. Wang, Z.~Wang, and J.~Guo, ``Dynamic datasets and market environments for financial reinforcement learning,'' 2023. [Online]. Available: \url{https://arxiv.org/abs/2304.13174}
\BIBentrySTDinterwordspacing

\bibitem{FinRL-DeepSeek}
\BIBentryALTinterwordspacing
M.~Benhenda, ``Finrl-deepseek: Llm-infused risk-sensitive reinforcement learning for trading agents,'' 2025. [Online]. Available: \url{https://arxiv.org/abs/2502.07393}
\BIBentrySTDinterwordspacing

\bibitem{reportfinancialregulationschallenge}
\BIBentryALTinterwordspacing
K.~Wang, J.~Patel, C.~Shen, D.~Kim, A.~Zhu, A.~Lin, L.~Borella, C.~Osborne, M.~White, S.~Yang, K.~Xiao, and X.-Y.~L. Yanglet, ``A report on financial regulations challenge at coling 2025,'' 2025. [Online]. Available: \url{https://arxiv.org/abs/2412.11159}
\BIBentrySTDinterwordspacing

\bibitem{FinMind-Y-Me}
\BIBentryALTinterwordspacing
P.~Chantangphol, P.~Balee, K.~Sucharitpongpan, C.~Saetia, and T.~Chalothorn, ``{F}in{M}ind-{Y}-me at the regulations challenge task: Financial mind your meaning based on {TH}a{LLE},'' in \emph{Proceedings of the Joint Workshop of the 9th Financial Technology and Natural Language Processing (FinNLP), the 6th Financial Narrative Processing (FNP), and the 1st Workshop on Large Language Models for Finance and Legal (LLMFinLegal)}, C.-C. Chen, A.~Moreno-Sandoval, J.~Huang, Q.~Xie, S.~Ananiadou, and H.-H. Chen, Eds.\hskip 1em plus 0.5em minus 0.4em\relax Abu Dhabi, UAE: Association for Computational Linguistics, Jan. 2025, pp. 349--362. [Online]. Available: \url{https://aclanthology.org/2025.finnlp-1.41/}
\BIBentrySTDinterwordspacing

\bibitem{simulatingfinancial2024}
\BIBentryALTinterwordspacing
S.~Gao, Y.~Wen, M.~Zhu, J.~Wei, Y.~Cheng, Q.~Zhang, and S.~Shang, ``Simulating financial market via large language model based agents,'' 2024. [Online]. Available: \url{https://arxiv.org/abs/2406.19966}
\BIBentrySTDinterwordspacing

\bibitem{econagent2024}
\BIBentryALTinterwordspacing
N.~Li, C.~Gao, M.~Li, Y.~Li, and Q.~Liao, ``{E}con{A}gent: Large language model-empowered agents for simulating macroeconomic activities,'' in \emph{Proceedings of the 62nd Annual Meeting of the Association for Computational Linguistics (Volume 1: Long Papers)}, L.-W. Ku, A.~Martins, and V.~Srikumar, Eds.\hskip 1em plus 0.5em minus 0.4em\relax Bangkok, Thailand: Association for Computational Linguistics, Aug. 2024, pp. 15\,523--15\,536. [Online]. Available: \url{https://aclanthology.org/2024.acl-long.829/}
\BIBentrySTDinterwordspacing

\bibitem{repeater2020}
\BIBentryALTinterwordspacing
A.~Holtzman, J.~Buys, L.~Du, M.~Forbes, and Y.~Choi, ``The curious case of neural text degeneration,'' 2020. [Online]. Available: \url{https://arxiv.org/abs/1904.09751}
\BIBentrySTDinterwordspacing

\bibitem{repeater2025}
\BIBentryALTinterwordspacing
Y.~Ren and D.~J. Sutherland, ``Learning dynamics of llm finetuning,'' 2025. [Online]. Available: \url{https://arxiv.org/abs/2407.10490}
\BIBentrySTDinterwordspacing

\bibitem{LlamaFactory2024}
\BIBentryALTinterwordspacing
Y.~Zheng, R.~Zhang, J.~Zhang, Y.~Ye, Z.~Luo, Z.~Feng, and Y.~Ma, ``Llamafactory: Unified efficient fine-tuning of 100+ language models,'' 2024. [Online]. Available: \url{https://arxiv.org/abs/2403.13372}
\BIBentrySTDinterwordspacing

\bibitem{DeepSpeed2023}
\BIBentryALTinterwordspacing
S.~A. Jacobs, M.~Tanaka, C.~Zhang, M.~Zhang, S.~L. Song, S.~Rajbhandari, and Y.~He, ``Deepspeed ulysses: System optimizations for enabling training of extreme long sequence transformer models,'' 2023. [Online]. Available: \url{https://arxiv.org/abs/2309.14509}
\BIBentrySTDinterwordspacing

\bibitem{openai2020}
\BIBentryALTinterwordspacing
J.~Kaplan, S.~McCandlish, T.~Henighan, T.~B. Brown, B.~Chess, R.~Child, S.~Gray, A.~Radford, J.~Wu, and D.~Amodei, ``Scaling laws for neural language models,'' 2020. [Online]. Available: \url{https://arxiv.org/abs/2001.08361}
\BIBentrySTDinterwordspacing

\bibitem{tsinghua2024}
\BIBentryALTinterwordspacing
Y.~Wu, Z.~Sun, S.~Li, S.~Welleck, and Y.~Yang, ``Inference scaling laws: An empirical analysis of compute-optimal inference for problem-solving with language models,'' 2024. [Online]. Available: \url{https://arxiv.org/abs/2408.00724}
\BIBentrySTDinterwordspacing

\bibitem{phasetransitions}
\BIBentryALTinterwordspacing
Y.~Sun and B.~Haghighat, ``Phase transitions in large language models and the $o(n)$ model,'' 2025. [Online]. Available: \url{https://arxiv.org/abs/2501.16241}
\BIBentrySTDinterwordspacing

\end{thebibliography}

\end{document}